\documentclass[10pt,twocolumn,letterpaper]{article}

\usepackage[pagenumbers]{cvpr} 

\usepackage[utf8]{inputenc}
\usepackage[T1]{fontenc}

\usepackage{amsmath, amsfonts}
\usepackage{nicefrac}           
\usepackage{siunitx}            
\usepackage{booktabs}           
\usepackage{multirow}
\usepackage{tabularx}
\usepackage{makecell}
\usepackage{colortbl}

\usepackage[dvipsnames]{xcolor} 

\usepackage[ruled]{algorithm2e}
\usepackage{algpseudocode}

\usepackage{microtype}
\usepackage{url}
\usepackage{enumitem}
\usepackage{float}

\usepackage{tikz}
\usetikzlibrary{arrows.meta, positioning}  

\usepackage{natbib}
\usepackage{float}
\usepackage{placeins}

\definecolor{cvprblue}{rgb}{0.21,0.49,0.74}
\usepackage[pagebackref,breaklinks,colorlinks,citecolor=cvprblue]{hyperref}

\def\paperID{306} 
\def\confName{3DV\xspace}
\def\confYear{2026\xspace}

\title{Next Best View Selections for Semantic and Dynamic 3D Gaussian Splatting}

\author{
Yiqian Li
\and Wen Jiang  \\
\\
University of Pennsylvania
\and Kostas Daniilidis
}

\begin{document}
\maketitle

\begin{abstract}
Understanding semantics and dynamics has been crucial for embodied agents in various tasks.
Both tasks have much more data redundancy than the static scene understanding task.
We formulate the view selection problem as an active learning problem, where the goal is to prioritize frames that provide the greatest information gain for model training. To this end, we propose an active learning algorithm with Fisher Information that quantifies the informativeness of candidate views with respect to both semantic Gaussian parameters and deformation networks. This formulation allows our method to jointly handle semantic reasoning and dynamic scene modeling, providing a principled alternative to heuristic or random strategies. We evaluate our method on large-scale static images and dynamic video datasets by selecting informative frames from multi-camera setups. Experimental results demonstrate that our approach consistently improves rendering quality and semantic segmentation performance, outperforming baseline methods based on random selection and uncertainty-based heuristics.
\end{abstract}    
\section{Introduction}
Recent advancements in 3D Gaussian Splatting (3DGS) have enabled real-time rendering and semantic scene understanding for both static and dynamic environments. Extensions of 3DGS now incorporate features from large-scale 2D vision-language models for semantic reasoning. To enable dynamic scene representations, recent methods~\cite{DNERF,dynerf,dynamic3dgs,4dgs,gs-flow,fully} augment Gaussians with temporal modeling to capture changes in geometry and appearance over time. These developments make 3DGS as a promising backbone for applications in robotics, AR/VR, and digital content creation, where efficiency and semantic accuracy are critical. However, the potential of 3DGS in such real-world settings is limited by the substantial resources required for training, especially when scaling to large-scale environments, densely instrumented multi-camera systems, or long-duration dynamic sequences. The key challenge lies in the vast number of candidate viewpoints: although many are redundant or uninformative, choosing which views are most informative is non-trivial. Unlike static reconstruction, dynamic environments introduce additional uncertainty due to evolving geometry, appearance changes, and semantic inconsistencies across time. 

A natural solution is \textbf{next-best-view} (NBV) selection, which seeks to prioritize views that most improve reconstruction and understanding. Effective NBV selection can significantly improve both geometric reconstruction and semantic consistency, especially in dynamic environments where scene content changes over time. In practice, however, existing strategies often rely on random sampling, uncertainty heuristics~\cite{CSS, activeneural}, or black-box selection models~\cite{dnrselect, activeinitsplat}, which do not explicitly maximize information gain and therefore can lead to suboptimal reconstruction and semantic coverage. Moreover, prior work predominantly focuses on static scenes, leaving dynamic NBV selection largely unexplored. Under dynamic scenes, these strategies struggle to generalize across scenes with complex motion and appearance variation, often lead to wasted computation, incomplete reconstructions, and semantic drift, leaving us without a principled way to allocate training resources effectively. 

In this work, we present a unified framework that combines dynamic and semantic 3DGS and formulates NBV selection as an active learning problem. Building on this formulation, we develop the first NBV algorithm tailored to dynamic semantic 3DGS, jointly capturing geometric, semantic, and temporal deformation information. Unlike FisherRF~\cite{fisherrf}, which considers only geometric information for static scenes, our method systematically leverages Fisher Information to identify views that maximally improve both reconstruction quality and dynamic scene modeling. To handle the computational challenges of large-scale dynamic semantic 3DGS, we propose an efficient diagonal approximation of the Fisher Information matrix and introduce a novel estimator for the Fisher Information of deformation networks using the trace of the gradient outer product. Beyond proposing a new algorithm, our framework integrates NBV selection directly into the 3DGS backbone, greatly extending the scenarios and applications of NBV to semantic and dynamic scene understanding. 

We evaluate our framework on large-scale static datasets (Replica~\cite{replica}) and dynamic multi-camera sequences (Neu3D~\cite{dynerf}), showing consistent improvements over baseline methods. Our contributions include:
\begin{itemize}
    \item The first Fisher Information-driven NBV selection framework for \emph{dynamic semantic 3DGS}, capturing geometric, photometric, semantic, and deformation informativeness.
    \item An efficient Fisher Information formulation for large-scale semantic and dynamic 3D Gaussian Splatting, which extends the diagonal approximation of FisherRF~\cite{fisherrf} to semantic Gaussian parameters and deformation networks, enabling tractable NBV selection beyond static radiance fields.
    \item A novel method to estimate Fisher Information for deformation MLPs, leveraging the trace of the gradient outer product as a proxy.
\end{itemize}

\section{Related works}
\subsection{Semantic 3D Gaussians Splatting}
Recent advancements have significantly improved semantic understanding in 3D scene representations. Early works build upon implicit radiance field representations such as NeRF. Methods like Semantic NeRF~\cite{semanticnerf}, Panoptic Lifting~\cite{Panopticlifting}, and Contrastive Lift~\cite{ContrastiveLift} pioneered the integration of semantic labels into NeRF-based frameworks, achieving sharp and accurate 3D semantic segmentation. Subsequent approaches—such as Distilled FFD~\cite{FFD}, F3F~\cite{F3F}, panoptic NeRF~\cite{panopticnerf}, NeRF-SOS~\cite{nerfsos}, Interactive Segmentation of Radiance Fields~\cite{ISRF}, Weakly Supervised 3D Open-Vocabulary Segmentation~\cite{weakly}, FFNF~\cite{FFNF}, VL-Fields~\cite{VL-Fields}, Featurenerf~\cite{featurenerf}, Distilled Feature Fields~\cite{DistilledFF}, and LERF~\cite{lerf} leverage semantic features from pretrained vision-language models like CLIP~\cite{CLIP}, LSeg\cite{Lseg}, or DINO\cite{dino} to enable open-vocabulary, pixel-level semantic understanding.

More recently, 3D Gaussian Splatting (3DGS)~\cite{3DGS} has emerged as a compelling explicit alternative to NeRF due to its faster training, real-time rendering capabilities, and high-quality scene reconstruction. Semantic extensions of 3DGS have shown great promise. Semantic Gaussians~\cite{semanticgs} and ConceptFusion~\cite{conceptfusion} project features from 2D pretrained encoders onto 3D Gaussians, enabling open-vocabulary understanding. OpenGaussian~\cite{open} and OpenScene~\cite{openscene} achieve point-level semantic segmentation and produce sharp and accurate semantic segmentation results. Feature 3DGS~\cite{feature3dgs}, CLIP-FO3D~\cite{Clip-fo3d}, and PLA~\cite{pla} further distill semantic features from 2D foundation models into the 3D domain using differentiable rasterization. Thanks to the real-time rendering and optimization capabilities of 3DGS, notable examples like SGS-SLAM~\cite{sgs}, SemGauss-SLAM~\cite{sem-slam}, and GS³LAM~\cite{gs3lam} also integrate semantic features into 3DGS-based SLAM pipelines for robust and scalable scene reconstruction.

Motivated by the strong generalization and ease of training of Feature 3DGS~\cite {feature3dgs}, we adopt its framework as the foundation for our semantic Gaussian Splatting module.

\subsection{Dynamic Gaussians Splatting}
Several recent works have extended 3D scene representation to dynamic scenes by introducing temporal modeling, enabling novel view synthesis in dynamic environments. Based on NeRF, early approaches such as D-NeRF~\cite{DNERF} introduced time as an explicit input to model non-rigid deformations, while Nerfies~\cite{nerfies} leveraged deformation fields with elastic regularization to animate casually captured scenes. HyperNeRF~\cite{HyperNeRF} advanced this by modeling topological changes via a hyperspace embedding, handling complex non-rigid transformations. STaR~\cite{star} decomposes scenes into static and dynamic components using a self-supervised rigid-body model. 

As for 3DGS-based models, 4D Gaussian Splatting (4D-GS)~\cite{4dgs} introduces a unified representation that combines 3D Gaussians with 4D neural voxels with a lightweight MLP to efficiently predict Gaussian deformations over time. Gaussian-Flow~\cite{gs-flow} alternatively models temporal attribute deformations using time and frequency domain parameterizations. Dynamic 3D Gaussians~\cite{dynamic3dgs} uses dense 6-DOF tracking and dynamic reconstruction through enforcing local rigidity constraints to enable Gaussians to maintain persistent attributes when they move and rotate over time. Explicit sampling strategies like Fully Explicit Dynamic Gaussian Splatting~\cite{fully} have also been proposed to separate static and dynamic components during training to represent continuous motion and reduce computational cost.

In our work, we adopt the 4D-GS~\cite{4dgs} framework as the foundation for our dynamic Gaussians module, leveraging its efficient modeling of dynamic scenes and real-time rendering capabilities.

\subsection{Uncertainty and view selection}
Driven by the need to reduce training costs and enable active training, a paradigm where models selectively acquire data based on informativeness, many recent works have focused on extracting useful cues from the trained model, existing data, and candidate viewpoints. A central theme across these methods is the estimation and exploitation of uncertainty. Magic Moments~\cite{MagicMoments} introduces higher moments of rendering equations to quantify the uncertainty of radiance fields and show their relationship with rendered error and their usage as powerful NBV selection criteria. CSS~\cite{CSS} utilizes the expectation and variance of categorical distributions to probabilistically update semantic maps within the 3D Gaussian Splatting (3DGS) framework. Similarly, ~\cite{activeneural} proposes a colorized surface voxel to interpret color uncertainty and surface information. 
While these methods primarily adopt Bayesian formulations to explicitly quantify uncertainty, others infer uncertainty by learning data distributions via alternative models. DNRSelect~\cite{dnrselect} employs a reinforcement learning-based view selector trained on rasterized images to find optimal views for deferred neural rendering. ActiveInitSplat~\cite{activeinitsplat} defines a black-box objective function based on density and occupancy metrics and employs a Gaussian process as the surrogate model to predict informative views.   

Beyond uncertainty estimation, several methods directly target maximizing information gain. GauSS-MI~\cite{gaussmi} utilizes Shannon Mutual Information to select informative viewpoints. FisherRF~\cite{fisherrf} introduces a Fisher Information-based metric to guide view selection in radiance fields without reliance on ground-truth data. AG-SLAM~\cite{agslam} further extends this idea by integrating it into an active SLAM framework, enabling online scene reconstruction and autonomous exploration via information-driven trajectory planning.

In this work, we adopt Fisher Information as the unified selection criterion for both semantic Gaussian Splatting and deformation-based dynamic modeling, facilitating view selection that balances scene semantics and motion-aware updates.

\section{Method}
In this section, we describe our active learning algorithm with Fisher Information for NBV selection in dynamic semantic 3DGS. We begin by reviewing the necessary preliminaries on semantic 3DGS and dynamic 3DGS, which form the foundation of our unified backbone. We then introduce our formulation for computing Fisher Information, first for Gaussian parameters, which capture both geometry and semantics features in Section 3.2 and subsequently for the deformation network, which captures temporal dynamics in Section 3.3. This decomposition allows us to quantify the informativeness of candidate views across all key aspects of the scene representation, enabling principled and efficient NBV selection.

\subsection{Preliminaries}
In this Section, we first introduce the two core components of our representation framework. Section 3.1.1 details the formulation of Semantic 3D Gaussian Splatting, which distills semantic features from 2D vision-language models into a 3D explicit representation. Section 3.1.2 extends this framework to 4D Gaussian Splatting, incorporating a deformation network to capture dynamic scene changes over time. 

\subsubsection{Semantic 3D Gaussians Splatting}

Following the work of feature 3DGS~\cite{feature3dgs}, the semantic 3D Gaussian Splatting (3DGS) represents a scene using a collection of Gaussians \(\{\mathcal{G}_i\}_{i=1}^N\), where each Gaussian \(\mathcal{G}_i\) is defined by its location \(x_i \in \mathbb{R}^3\), rotation \(q_i \in \mathbb{R}^4\), scale \(s_i \in \mathbb{R}^3\), opacity \(\alpha_i \in \mathbb{R}\), color \(c_i \in \mathbb{R}^3\), and a semantic feature vector \(f_i \in \mathbb{R}^d\). The full parameter set is:
\[
\Theta_i = \{x_i, q_i, s_i, \alpha_i, c_i, f_i\}.
\]

To project the 3D Gaussians into image space, each covariance matrix \(\Sigma_i\) is transformed to camera space:
\begin{equation}
\Sigma'_i = J W \Sigma_i W^\top J^\top,
\end{equation}
where \(W\) is the world-to-camera transformation and \(J\) is the Jacobian of the projection function. The original 3D covariance is decomposed as:
\begin{equation}
\Sigma_i = R_i S_i^2 R_i^\top,
\end{equation}
where \(R_i\) is the rotation matrix, which can be converted from quaternions \(q_i\)) and \(S_i\) is a diagonal matrix of scales \(s_i\). In this way, the \(\Sigma_i\) can be guaranteed to be positive semi-definite during optimization.

Rendering is performed using volumetric \(\alpha\)-compositing in front-to-back order. The color \(C\) and semantic feature \(F_s\) at a pixel are computed as:
\begin{equation}
C = \sum_{i \in \mathcal{N}} c_i \alpha_i T_i, \quad
F_s = \sum_{i \in \mathcal{N}} f_i \alpha_i T_i,
\end{equation}
where \(\mathcal{N}\) is the set of Gaussians overlapping the pixel (sorted by depth), and the transmittance \(T_i\) is:
\begin{equation}
T_i = \prod_{j=1}^{i-1} (1 - \alpha_j).
\end{equation}

In order to supervise the semantic features, we distill from the pretrained 2D foundation model Lseg~\cite{Lseg} with a feature dimension of 512. To reduce training and rendering cost, we keep low-dimensional features \(f_i \in \mathbb{R}^{128}\) at each 3D Gaussian and employ a lightweight convolutional decoder to upsample the rendered feature map \(F_s \in \mathbb{R}^{H \times W \times 128}\) to \(F_s' \in \mathbb{R}^{H \times W \times 512}\).  The overall loss function is:
\begin{equation}
\mathcal{L} = \mathcal{L}_{\text{rgb}} + \gamma \mathcal{L}_{\text{feat}},
\end{equation}
with:
\[
\mathcal{L}_{\text{rgb}} = (1 - \lambda)\|I - \hat{I}\|_1 + \lambda \mathcal{L}_{\text{D-SSIM}}, \quad
\mathcal{L}_{\text{feat}} = \|F_t - F_s'\|_1,
\]
where \(\hat{I}\) is the rendered image and \(I\) is the ground truth.

\subsubsection{Dynamic 3DGS}

4D-GS~\cite{4dgs} extends the static 3D Gaussian Splatting framework to dynamic scenes by introducing a deformation network that models the motion and shape changes of 3D Gaussians over time. At any timestamp \(t\), the Gaussians are transformed via a deformation network:
\begin{equation}
G'_t = G + \Delta G_t = \{X_i + \Delta X_i(t), s_i + \Delta s_i(t), r_i + \Delta r_i(t)\},
\end{equation}
where \(\Delta X_i(t)\), \(\Delta s_i(t)\), and \(\Delta r_i(t)\) are the time-dependent predicted deformations.

The deformation network \(F_{\phi}\) consists of two components: a spatial-temporal structure encoder \(H\) and a multi-head deformation decoder \(D\). The encoder \(H\) uses a multi-resolution voxel grid decomposition, inspired by K-Planes~\cite{Kp}, over six 2D plane projections \((x,y), (x,z), (y,z), (x,t), (y,t), (z,t)\). Each Gaussian’s position and time \((X_i, t)\) are queried from these planes via bilinear interpolation to produce a feature embedding:
\begin{equation}
F_{i} = \phi_d\left(\bigcup_{(i,j)} \text{interp}(R_{(i,j)}^{(l)})\right),
\end{equation}
where \(R_{(i,j)}^{(l)}\) denotes the voxel planes at resolution level \(l\), and \(\phi_d\) is a small MLP used to fuse the plane features.

The decoder \(D = \{\phi_x, \phi_r, \phi_s\}\) consists of three heads that output the deformation for position, rotation, and scale:
\begin{equation}
\Delta X_i(t) = \phi_x(F_i), \quad \Delta r_i(t) = \phi_r(F_i), \quad \Delta s_i(t) = \phi_s(F_i).
\end{equation}

During rendering, the pretrained 3DGS are first deformed through $\{G'_i\}_{i=1}^N = F\left(\{G_i\}_{i=1}^N\right)$. Then, the deformed Gaussians \(G'_t\) are rendered via differentiable splatting to produce the final image:
\begin{equation}
\hat{y}_t = S(M_t, G'_t),
\end{equation}
where \(M_t\) is the camera pose at time \(t\) and \(S\) denotes the splatting renderer. 

\begin{figure*}[t]
  \centering
  \includegraphics[width=0.95\textwidth]{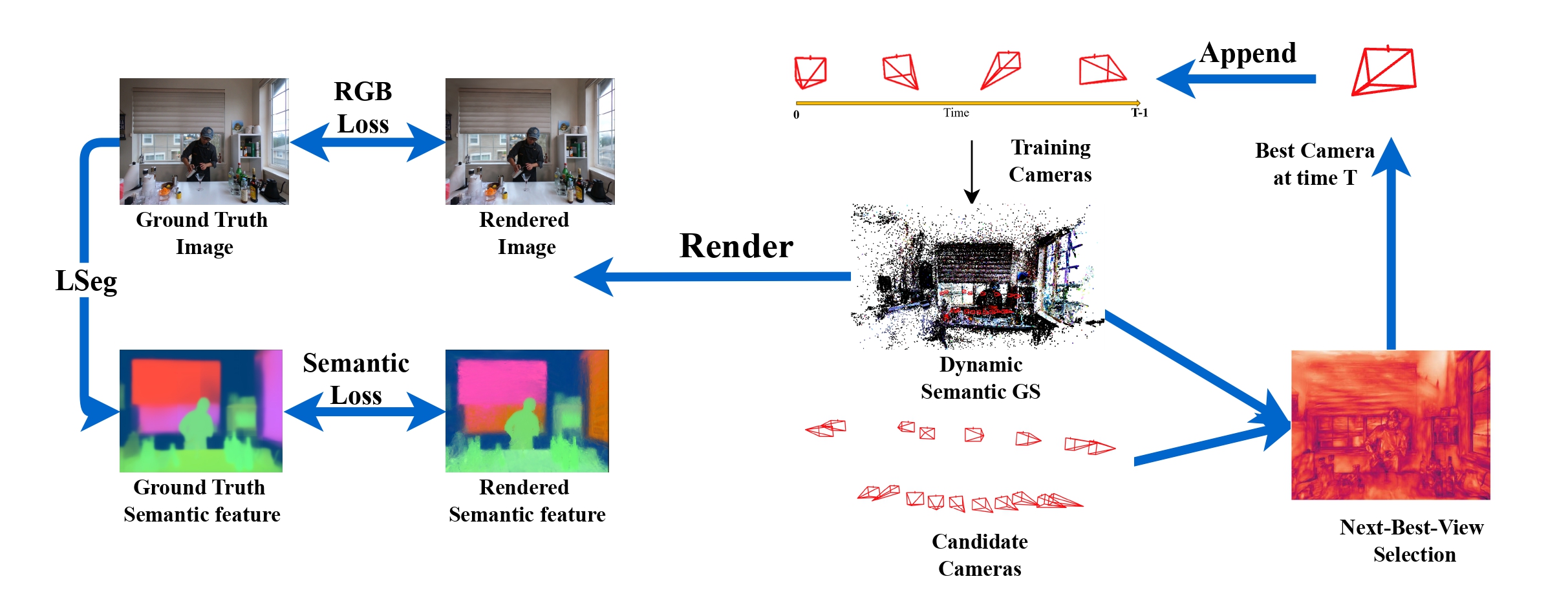}
  \caption{\textbf{Overview of our work.} At each timestep, we select the NBV from a set of candidate camera views using a Fisher Information-based criterion. The selection process evaluates each candidate view based on its expected contribution to both semantic feature learning and dynamic deformation modeling. Once selected, the chosen view is incorporated into the training process of our Dynamic and Semantic 3DGS. For rendering and supervision, the Gaussians are first deformed by an MLP deformation network based on the current timestep, then projected into RGB images and semantic feature maps. These outputs are compared against ground-truth images and semantic labels to jointly optimize both the dynamic deformation and semantic representation.}
  \label{fig:pipeline}
\end{figure*}

\subsection{Fisher information for semantic 3DGS}
In this section, we will describe how to qualify Fisher information for Gaussian parameters for NBV selection.\\
Following the instruction of FisherRF~\cite{fisherrf}, let the radiance field be parameterized by \( G \), representing the set of attributes for all 3D Gaussians in the scene, including location \( x \), scale \( s \), opacities \( \alpha \), colors \( c \), and semantic feature \( f \). At a given camera pose \( M \in \text{SE}(3) \), the rendering function \( S(M, G) \) produces a synthetic image. We consider the negative log-likelihood of the observed image \( y \) under the rendered prediction:
\begin{equation}
\mathcal{L}(y; G) = -\log p(y | M,G) = \| y -  S(M, G) \|^2_2.
\end{equation} 
Under standard regularity conditions, the \emph{observed} Fisher Information with respect to the parameters \( G \) at pose \( M \) is given by the Hessian of this loss:
\begin{equation}
\mathcal{I}(G; M) = \nabla_G  S(M, G)^\top \nabla_G  S(M, G) + \lambda I,
\label{eq:fisherreg}
\end{equation}
where \( \lambda I \) is a log-prior regularization term.

However, due to the large number of parameters, computing the full Fisher Information matrix is intractable in practice. Following the approximation proposed in FisherRF~\cite{fisherrf}, we apply a Laplace approximation by retaining only the diagonal elements of the Hessian and adding a prior regularization term:

\begin{equation}
\mathcal{I}(G; M) \approx \text{diag}\left( \nabla_G S(M, G)^\top \nabla_G S(M, G) \right) + \lambda I,
\end{equation}
 This diagonal approximation significantly reduces computational cost and memory overhead, while preserving the essential structure of the Fisher Information needed for evaluating local sensitivity.

To select the most informative viewpoint from a candidate pool \( \{M_i\}_{i=1}^N \), we pick the view that maximize the \emph{Expected Information Gain} (EIG), defined as:
\begin{equation}
\text{EIG}(M_i) = \mathrm{tr}\left( \mathcal{I}(G; M_i) \cdot \mathcal{I}_{\text{train}}^{-1} \right),
\end{equation}
where \( \mathcal{I}_{\text{train}} \) is the accumulated Fisher Information of current training views. The trace measures how much the candidate view would sharpen the current uncertainty landscape over parameters.

\subsection{Fisher Information for Deformation Network}
In this section, we introduce a novel estimator for the Fisher Information of multilayer perceptrons (MLPs), which we employ to quantify the informativeness of our deformation module.

For the deformation network \(F_{\phi}\), the informativeness of a candidate view \((\mathbf{x}^{acq}, \mathbf{y}^{acq})\) is quantified by the Fisher Information with respect to deformation network parameters \(\phi\) and the current training data \(D^{train}\):
\begin{align}
\mathcal{I}[\phi ; \mathbf{y}^{acq} \mid \mathbf{x}^{acq}, D^{train}] 
&= H[\phi \mid D^{train}] \nonumber \\
&\quad - H[\phi \mid \mathbf{y}^{acq}, \mathbf{x}^{acq}, D^{train}].
\end{align}

Following FisherRF~\cite{fisherrf}, this gain can be upper bounded by the curvature of the log-likelihood:
\begin{align}
\mathcal{I}[\phi] 
&\;\le\; \tfrac{1}{2}\,\operatorname{tr} \Big(
    \mathbf{H}''_{\phi}\big[ \log p(\mathbf{y}^{acq} \mid \mathbf{x}^{acq}, \phi) \big] 
    \nonumber \\
&\quad \cdot \;
    \mathbf{H}''_{\phi}\big[ \log p(\phi \mid D^{train}) \big]^{-1}
\Big).
\end{align}

where \(\mathbf{H}''_{\phi}\) denotes the Hessian with respect to \(\phi\).

For view selection, the prior Hessian term \(\mathbf{H}''_{\phi}[\log p(\phi \mid D^{train})]^{-1}\) is constant across candidate views and can be dropped. Thus our acquisition criterion simplifies to maximizing
\[
\mathcal{S}(\mathbf{x}^{acq}) 
= \operatorname{tr}\!\left( \mathbf{H}''_{\phi}\big[ \log p(\mathbf{y}^{acq} \mid \mathbf{x}^{acq}, \phi) \big] \right).
\]

Finally, we estimate the trace efficiently with Hutchinson’s estimator:
\[
\mathcal{S}(\mathbf{x}^{acq}) 
\;\approx\; \mathbb{E}_{\mathbf{v} \sim \mathcal{N}(0, I)} \big[ \| \mathbf{H}''_{\phi}\, \mathbf{v} \|^2 \big],
\]
where \(\mathbf{v}\) is a random probe vector and Hessian-vector products are computed via automatic differentiation.

This formulation explicitly ties Fisher Information to the deformation network parameters \(\phi\), ensuring that the selected views maximize temporal consistency and improve the modeling of dynamic scene evolution.

\section{Experiments}
In this section, we present a comprehensive evaluation of our proposed method. We first describe the experimental setup, including the datasets, baseline methods, and evaluation metrics. We then report results on two key scenarios: selecting a limited number of candidate camera poses from large-scale semantic datasets for static semantic 3DGS, and selecting informative frames from multi-camera sequences for dynamic and semantic 3DGS. Finally, we conduct ablation studies to analyze the impact of the log-prior regularization parameter and different feature choices for NBV selection, providing deeper insights into the design and effectiveness of our approach.

\subsection{Experimental Set-up}
\paragraph{Dataset}
For semantic 3DGS active training, we choose Replica~\cite{replica}, which composes various indoor environments with semantic labels. We follow the dataset setting of Feature 3DGS ~\cite{feature3dgs} to pick 75-80 images for each scene and leave 15 of them for testing. \\
For dynamic and semantic 3DGS active training, we choose Neu3D~\cite{dynerf}, which includes 15-21 synchronous camera videos. From each view, we sample 40 temporally synchronized frames. The central view, \texttt{cam00}, is reserved as the test view, and the remaining views are used for training. \\ 
To gain the ground truth semantic features, we apply LSeg~\cite{Lseg} to semantic segmentation and attain pixel-level features for each image. We utilize Colmap~\cite{schoenberger2016mvs}~\cite{schoenberger2016sfm} to get sparse point clouds for training initialization. 

\paragraph{Baselines}
We compare our method against random view selection and semantic uncertainty-based selection. Following the approach of Magic Moments~\cite{MagicMoments}, we use feature covariance as a measure of uncertainty and implement the computation efficiently in CUDA. To estimate the second moment of semantic features, we first square each feature vector associated with a Gaussian and accumulate the results to compute the per-pixel second moment. The covariance is then obtained by subtracting the squared mean (i.e., the square of the rendered semantic output) from the second moment.

\paragraph{Metrics}
We evaluate our rendering quality with Structural Similarity Index Measure (SSIM)~\cite{ssim}, Peak-signal-to-noise ratio (PSNR), and Learned Perceptual Image Patch Similarity (LPIPS)~\cite{lpips}, and semantic segmentation quality with mean intersection-over-union (mIoU) and mean accuracy (mAcc).  
To evaluate semantic segmentation, we first project Gaussian semantic features into rendered images to get per-pixel features. Then, we project feature maps into class probabilities by computing dot products between features and text features from predefined labels and assigning every pixel to the class with the highest logit per pixel.

 \subsection{Static semantic 3DGS active training}
For active training in static semantic Gaussian Splatting, we first start our training with two images. Then after 200 iterations, we compute Fisher Information for all the remained camera poses and add the one that yields the highest expected information gain to the training set, until adding 10 images every 200 iterations. The model is then trained for a total of 7,000 iterations. All experiments are conducted on an NVIDIA L40 GPU, requiring approximately 4GB of memory and 45 minutes of training per scene. For each NBV selection, it takes approximately 7 seconds.

Table~\ref{tab:replica_results} summarizes the average performance of different view selection strategies on the Replica dataset. Our method, which incorporates Fisher Information computed with semantic information, achieves the best results across all metrics, indicating improved visual quality and semantic accuracy.

Removing the semantic information from the Fisher Information calculation leads to a consistent decline in performance, highlighting the importance of incorporating semantic sensitivity when estimating view informativeness. The semantic covariance baseline Magic Moments ~\cite{MagicMoments}, which relies on second-order statistics of categorical features, underperforms compared to our Fisher Information-based method with semantic information. The random view selection baseline results in the weakest performance across all metrics, further emphasizing the advantage of principled, information-driven view selection.

\begin{figure*}[h!]
  \centering
  \includegraphics[width=0.9\linewidth]{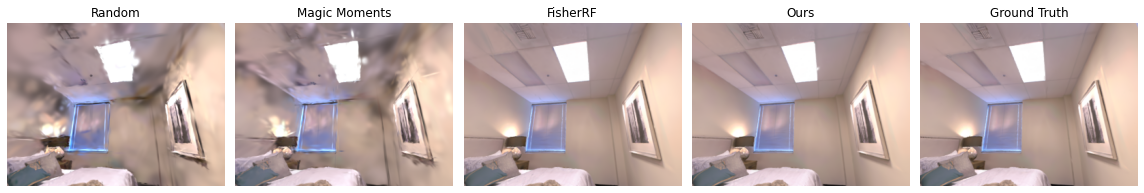}
  \includegraphics[width=0.9\linewidth]{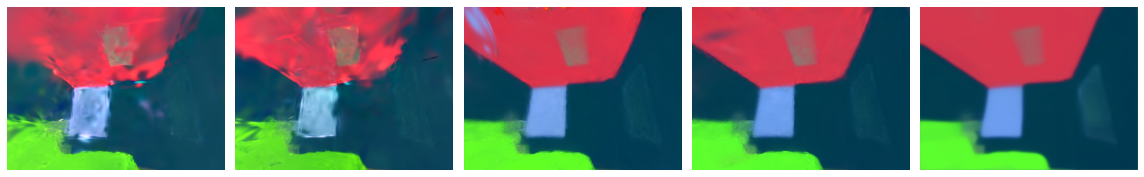}
  \includegraphics[width=0.9\linewidth]{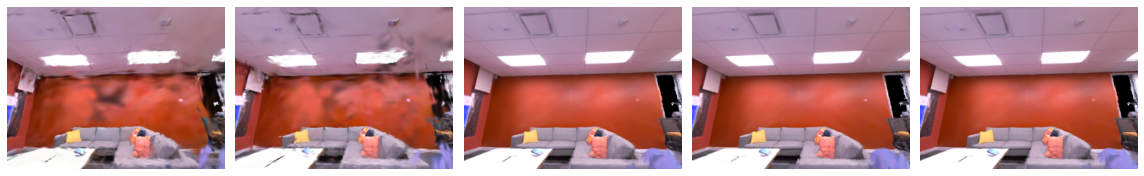}
  \includegraphics[width=0.9\linewidth]{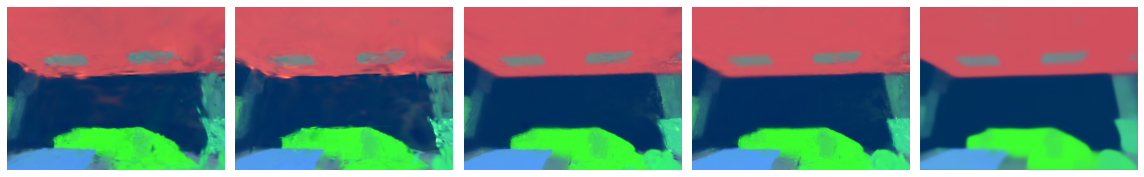}
  \caption{\textbf{Semantic segmentation results on Replica dataset.} The first and third rows of images are the rendered results. The second and forth rows of images are the results of visualizing the feature maps of test set after being trained with 12 training views. All the methods are based on Feature 3DGS~\cite{feature3dgs} except for different view selection methods to augment the training data.}
\end{figure*}

\definecolor{tabfirst}{rgb}{1, 0.7, 0.7} 
\definecolor{tabsecond}{rgb}{1, 0.85, 0.7} 
\definecolor{tabthird}{rgb}{1, 1, 0.7} 

\begin{table}[t]
\centering
\small 
\setlength{\tabcolsep}{4pt} 
\resizebox{\linewidth}{!}{%
    \begin{tabular}{lccccc}
    \toprule
    Method & SSIM (↑) & PSNR (↑) & LPIPS (↓) & mAcc (↑) & mIoU (↑) \\
    \midrule

Random                             &                      0.9077 &                      29.6600 &                      0.1477  &                      0.9189 &                      0.7120 \\
Magic Moments ~\cite{MagicMoments}                & \cellcolor{tabsecond}0.9615 & \cellcolor{tabsecond}33.5200 & \cellcolor{tabsecond}0.0591  & \cellcolor{tabsecond}0.9256 & \cellcolor{tabsecond}0.7593 \\
FisherRF ~\cite{fisherrf}  &  \cellcolor{tabthird}0.9593 &  \cellcolor{tabthird}32.1700 &  \cellcolor{tabthird}0.0614  &  \cellcolor{tabthird}0.9230 &  \cellcolor{tabthird}0.7441 \\
Ours &  \cellcolor{tabfirst}0.9626 &  \cellcolor{tabfirst}33.6000 &  \cellcolor{tabfirst}0.0572  &  \cellcolor{tabfirst}0.9289 &  \cellcolor{tabfirst}0.7648\\
    \bottomrule

    \end{tabular}
    }
    \caption{Average performance on Replica dataset under different view selection strategies. The \colorbox{tabfirst}{\textbf{best}}, \colorbox{tabsecond}{\textbf{second best}}, and \colorbox{tabthird}{\textbf{third best}} results are highlighted with light red, orange, and yellow backgrounds, respectively.}
    \label{tab:replica_results}
\end{table}

 \subsection{Dynamic and semantic 3DGS active training}
 For active training in dynamic and semantic 3DGS, we begin by training a static semantic model using the first frame from all cameras for 3000 iterations. Subsequently, to train the deformation network for dynamic, for each future timestep, we compute the Fisher Information across all candidate camera views and iteratively select the view with the highest information gain every 200 iterations. This process continues until all 39 remaining timesteps are each assigned a single informative camera frame. The dynamic training phase runs 14000 iterations. For regularization, here we set \( \lambda = 10^{-6} \). Training is conducted on an NVIDIA L40 GPU, requiring approximately 8GB of memory and 2 hours per scene. For each NBV selection, it takes approximately 20 seconds.

 Table~\ref{tab:dynerf_results} presents the average performance across five dynamic scenes from the Neu3D dataset using different view selection strategies. The proposed method based on Fisher Information achieves the best performance across all evaluation metrics, presenting superior visual quality and semantic consistency. In contrast, the semantic covariance baseline consistently underperforms the Fisher-based method. The random selection baseline yields the weakest performance, especially in perceptual quality and segmentation accuracy, further validating the effectiveness of Fisher Information-guided view selection for dynamic and semantic 3DGS.

\begin{figure*}
  \centering
  \includegraphics[width=0.9\linewidth]{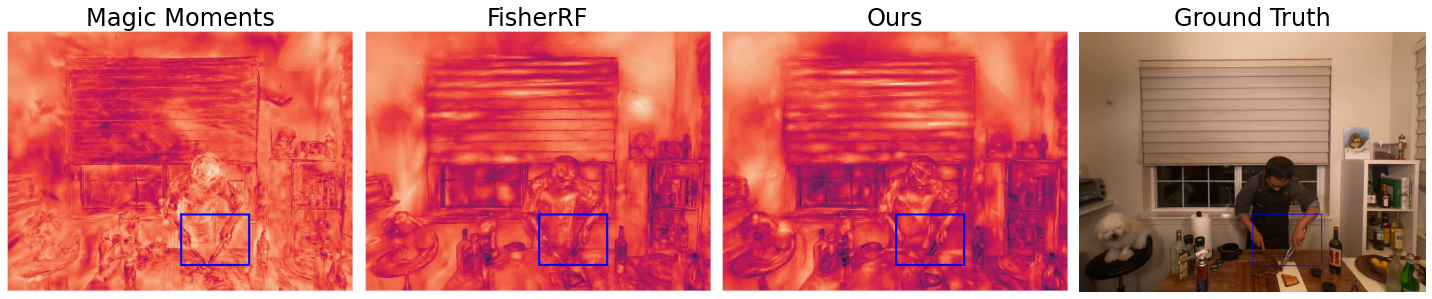}
  \includegraphics[width=0.9\linewidth]{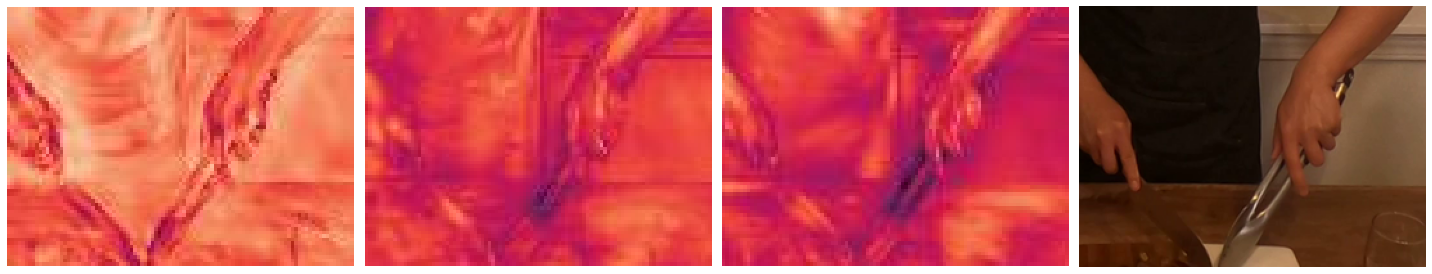}
  \caption{\textbf{Heatmap for selection method interest on Neu3D dataset.} The second row is zoom-in figures.}
  \label{heatmap}
\end{figure*}

We also analyze the attention patterns of the three selection methods. As shown in Fig.~\ref{heatmap}, which visualizes the interest heatmaps, the subject is in the process of closing the kitchen tongs. Magic Moments~\cite{MagicMoments} distributes attention evenly across the entire kitchen tongs. FisherRF~\cite{fisherrf} emphasizes the connection between the handles and the gripping ends. In contrast, our method focuses more on the handles, likely due to their dynamic motion during the closing action.

\begin{figure*}
  \centering
  \includegraphics[width=0.9\linewidth]{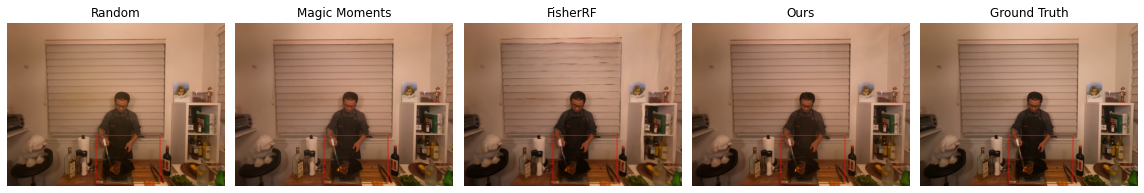}
  \includegraphics[width=0.9\linewidth]{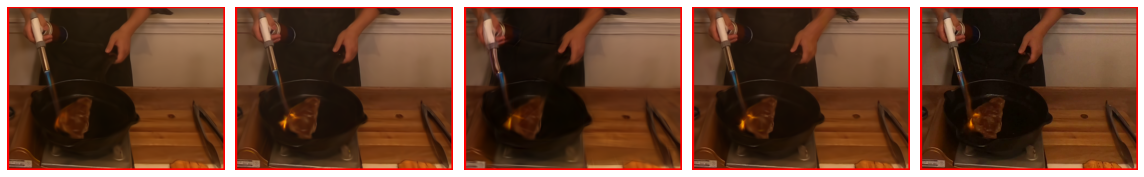}
  \includegraphics[width=0.9\linewidth]{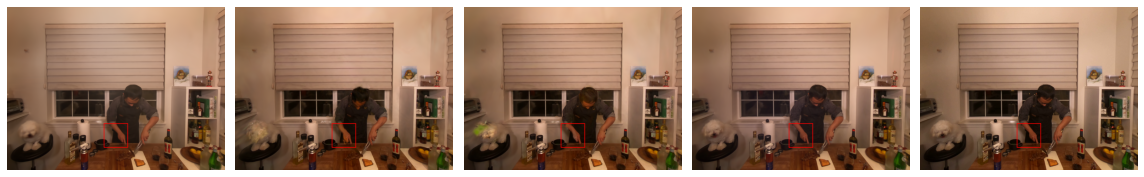}
  \includegraphics[width=0.9\linewidth]{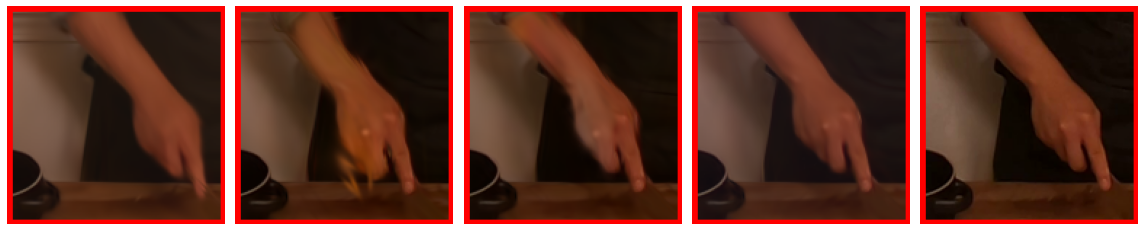}
  \caption{\textbf{Zoomed-in qualitative study of our method on Neu3D dataset.} The second and fourth rows are zoom-in figures. Visualizations are the results of the test set after being trained with 39 training views. All the methods are based on the same dynamic and semantic 3DGS, except for different view selection methods to augment the training data.}
\end{figure*}

\begin{table}
\centering
\resizebox{\columnwidth}{!}{
\begin{tabular}{lccccc}
\toprule
Method & SSIM (↑) & PSNR (↑) & LPIPS (↓) & mAcc (↑) & mIoU (↑) \\
\midrule
Random & 0.8831 & 26.7361 & 0.1933 & 0.8848 & 0.7284 \\
Magic Moments~\cite{MagicMoments} & \cellcolor{tabthird}0.9184 & \cellcolor{tabsecond}28.3795 & \cellcolor{tabsecond}0.1581 & \cellcolor{tabsecond}0.8890 & \cellcolor{tabthird}0.7350 \\
FisherRF~\cite{fisherrf} & \cellcolor{tabsecond}0.9186 & \cellcolor{tabthird}27.6743 & \cellcolor{tabthird}0.1605 & \cellcolor{tabthird}0.8863 & \cellcolor{tabsecond}0.7547 \\
Ours & \cellcolor{tabfirst}0.9239 & \cellcolor{tabfirst}28.6191 & \cellcolor{tabfirst}0.1553 & \cellcolor{tabfirst}0.8963 & \cellcolor{tabfirst}0.7604 \\
\bottomrule
\end{tabular}
}
\caption{Average performance across Neu3D under different view selection strategies.}
\label{tab:dynerf_results}
\end{table}

\subsection{Ablation study}


\begin{table}[t]
\centering
\small 
\setlength{\tabcolsep}{4pt} 
\resizebox{\linewidth}{!}{%
\begin{tabular}{lccccc}
\toprule
Method & SSIM (↑) & PSNR (↑) & LPIPS (↓) & mAcc (↑) & mIoU (↑) \\
\midrule
Ours (w/o reg.) & 0.9173 & \cellcolor{tabsecond}28.5853 & \cellcolor{tabthird}0.1589 & \cellcolor{tabsecond}0.8929 & \cellcolor{tabsecond}0.7413 \\
Ours ($\lambda = 10^{-7}$) & \cellcolor{tabthird}0.9184 & 28.0638 & 0.1594 & 0.8881 & \cellcolor{tabthird}0.7387 \\
Ours ($\lambda = 10^{-6}$) & \cellcolor{tabfirst}0.9239 & \cellcolor{tabfirst}28.6191 & \cellcolor{tabfirst}0.1553 & \cellcolor{tabfirst}0.8963 & \cellcolor{tabfirst}0.7604 \\
Ours ($\lambda = 10^{-5}$) & \cellcolor{tabsecond}0.9186 & \cellcolor{tabthird}28.5295 & \cellcolor{tabsecond}0.1584 & \cellcolor{tabthird}0.8898 & 0.7374 \\
\bottomrule
\end{tabular}%
}
\caption{Average performance on Neu3D dataset under our view selection strategy with and without regularization.}
\label{tab:Ablative}
\end{table}

\begin{table}
\centering
\resizebox{\columnwidth}{!}{
\begin{tabular}{lccccc}
\toprule
Method & SSIM (↑) & PSNR (↑) & LPIPS (↓) & mAcc (↑) & mIoU (↑) \\
\midrule
Geom~\cite{fisherrf} & 0.9186 & 27.6743 & 0.1605 & 0.8863 & \cellcolor{tabthird}0.7547 \\
Geom+Sem & 
\cellcolor{tabsecond}0.9200 & \cellcolor{tabsecond}28.2788 & \cellcolor{tabthird}0.1596 & \cellcolor{tabthird}0.8885 & \cellcolor{tabsecond}0.7571 \\
Geo+Def & 
\cellcolor{tabthird}0.9198 & \cellcolor{tabthird}28.1501 & \cellcolor{tabsecond}0.1578 & \cellcolor{tabsecond}0.8905 & 0.7367 \\
Ours & \cellcolor{tabfirst}0.9239 & \cellcolor{tabfirst}28.6191 & \cellcolor{tabfirst}0.1553 & \cellcolor{tabfirst}0.8963 & \cellcolor{tabfirst}0.7604 \\
\bottomrule
\end{tabular}
}
\caption{Average performance across Neu3D under different view selection strategies.}
\label{tab:Ablative2}
\end{table}

\noindent\textbf{Ablation on choice of log-prior regularization parameter.} 
We conduct an ablation study to evaluate the effect of the log-prior regularization parameter \( \lambda \) in the Fisher Information computation (Eq.~\ref{eq:fisherreg}). This term is critical for stabilizing the diagonal approximation of the Hessian, especially in dynamic and semantic scenes. Table~\ref{tab:Ablative} compares four variants of our method on the Neu3D dataset: no regularization (\( \lambda = 0 \)), weak regularization (\( \lambda = 10^{-7} \)), and our default setting (\( \lambda = 10^{-6} \)), and  strong regularization (\( \lambda = 10^{-5} \)). The results show that the full model with \( \lambda = 10^{-6} \) consistently outperforms the other three across all metrics. Without regularization, the Fisher scores become noisier and less reliable, leading to degraded view selection and semantic accuracy. 

\noindent\textbf{Ablation on Features Used for NBV Selection.} 
We conduct an ablation study to evaluate the contribution of different features in the Fisher Information computation for NBV selection in 4DGS. Specifically, we compare four variants on the Neu3D dataset: 
\textbf{Geom}, which uses Fisher information only on geometric parameters following \textit{FisherRF}~\cite{fisherrf}; 
\textbf{Geom+Sem}, which additionally incorporates semantic features; 
\textbf{Geom+Def}, which combines geometric and deformation parameters; and 
\textbf{Ours}, which jointly considers geometric, semantic, and deformation parameters. 
As shown in Table~\ref{tab:Ablative2}, adding semantic information improves fine-detail reconstruction and high-frequency texture fidelity, while adding deformation information enhances temporal consistency in dynamic regions. 
Our full model, which integrates all three components, achieves the best performance across all metrics, confirming that semantic and deformation cues provide complementary benefits to purely geometric NBV selection.

\section{Conclusion}
To address the critical challenge of reducing training data requirements while preserving both rendering fidelity and semantic accuracy across static and dynamic scenes, we introduce a unified 3DGS backbone that jointly models semantic and temporal representations, together with an active learning framework based on Fisher Information for Next-Best-View selection. By quantifying the informativeness of candidate viewpoints with respect to geometric and semantic Gaussian parameters and the deformation network, our method identifies the most informative viewpoints to guide the training process. Our experimental results on Replica and Neu3D validate the effectiveness of the approach. These findings highlight that view selection guided by explicit information measures can lead not only to faster convergence but also to more reliable modeling of dynamic and semantic scenes. By reducing the reliance on large amounts of training data, our method enables more efficient and scalable training of high-quality 3DGS models.\\
Our framework offers a foundation for adaptive data acquisition. In robotics and SLAM, principled NBV selection can help agents actively choose viewpoints to improve mapping accuracy and semantic awareness under limited sensing budgets. In AR/VR, efficient view selection accelerates scene capture and dynamic updates. These impacts highlight the potential of our approach to enable efficient, scalable, and intelligent scene understanding across diverse domains.

\paragraph{Acknowledgment}
We greatly acknowledge financial support by the grants NSF FRR
2220868, NSF IIS-RI 2212433, and ONR N00014-22-1-2677.

\FloatBarrier
{
    \small
    \bibliographystyle{ieeenat_fullname}
    \bibliography{references}
}


\end{document}